\documentclass[english]{article}
\usepackage[latin9]{inputenc}
\usepackage{geometry}
\usepackage{array}
\usepackage{float}
\usepackage{amssymb}
\usepackage{graphicx}
\usepackage{amsbsy}
\usepackage{babel}

\providecommand{\tabularnewline}{\\}

\floatstyle{ruled}
\newfloat{algorithm}{tbp}{loa}
\providecommand{\algorithmname}{Algorithm}
\floatname{algorithm}{\protect\algorithmname}

\begin{document}

\title{Learning From Ordered Sets and Applications in Collaborative Ranking}
\author{Truyen Tran$^{\dagger\ddagger}$, Dinh Phung$^{\dagger}$, Svetha Venkatesh$^{\dagger}$ \\
   		$^{\dagger}$Pattern Recognition and Data Analytics, Deakin University, Australia\\
   		$^{\ddagger}$Department of Computing, Curtin University, Australia\\
   		\{truyen.tran,dinh.phung,svetha.venkatesh\}@deakin.edu.au
}

\date{}

\maketitle
\begin{abstract}
Ranking over sets arise when users choose between groups of items.
For example, a group may be of those movies deemed $5$ stars to them,
or a customized tour package. It turns out, to model this data type
properly, we need to investigate the general combinatorics problem
of partitioning a set and ordering the subsets. Here we construct
a probabilistic log-linear model over a set of ordered subsets. Inference
in this combinatorial space is highly challenging: The space size
approaches $(N!/2)6.93145^{N+1}$ as $N$ approaches infinity. We
propose a \texttt{split-and-merge} Metropolis-Hastings procedure that
can explore the state-space efficiently. For discovering hidden aspects
in the data, we enrich the model with latent binary variables so that
the posteriors can be efficiently evaluated. Finally, we evaluate
the proposed model on large-scale collaborative filtering tasks and
demonstrate that it is competitive against state-of-the-art methods.
\end{abstract}


\global\long\def\objects{\mathit{X}}

\global\long\def\bpi{\boldsymbol{\pi}}

\global\long\def\bsigma{\boldsymbol{\sigma}}

\global\long\def\rating{r}

\global\long\def\Stirling{S}

\global\long\def\logll{\mathcal{L}}

\global\long\def\remainset{R}

\global\long\def\hidden{\boldsymbol{h}}

\global\long\def\state{\boldsymbol{x}}

\global\long\def\ll{\mathcal{L}}

\global\long\def\Real{\mathbb{R}}

\global\long\def\genModel{\mbox{OSM}}

\section{Introduction}

Rank data has recently generated a considerable interest within the
machine learning community, as evidenced in ranking labels \cite{dekel2003llm,vembu2010label}
and ranking data instances \cite{cohen1999learning,weimer2008cofi}.
The problem is often cast as generating a list of objects (e.g., labels,
documents) which are arranged in decreasing order of relevance with
respect to some query (e.g., input features, keywords). The treatment
effectively ignores the grouping property of compatible objects \cite{wagstaff2010modelling}.
This phenomenon occurs when some objects are likely to be grouped
with some others in certain ways. For example, a grocery basket is
likely to contain a variety of goods which are complementary for household
needs and at the same time, satisfy weekly budget constraints. Likewise,
a set of movies are likely to given the same quality rating according
to a particular user. In these situations, it is better to consider
ranking groups instead of individual objects. It is beneficial not
only when we need to recommend a subset (as in the case of grocery
shopping), but also when we just want to produce a ranked list (as
in the case of watching movies) because we would better exploit the
compatibility among grouped items.

This poses a question of how to group individual objects into subsets
given a list of all possible objects. Unlike the situation when the
subsets are pre-defined and fixed (e.g., sport teams in a particular
season), here we need to explore the space of set partitioning and
ordering simultaneously. In the grocery example we need to partition
the stocks in the store into baskets and then rank them with respect
to their utilities; and in the movie rating example we group movies
in the same quality-package and then rank these groups according to
their given ratings. The situation is somewhat related to multilabel
learning, where our goal is to produce
a subset of labels out of many for a given input, but it is inherently
more complicated: not only we need to produce all subsets, but also
to rank them.

This paper introduces a probabilistic model for this type of situations,
i.e., we want to learn the statistical patterns from which a set of
objects is partitioned and ordered, and to compute the probability
of any scheme of partitioning and ordering. In particular, the model
imposes a log-linear distribution over the joint events of partitioning
and ordering. It turns out, however, that the state-space is prohibitively
large: If the space of complete ranking has the complexity of $N!$
for $N$ objects, then the space of partitioning a set and ordering
approaches $(N!/2)6.93145^{N+1}$ in size as $N$ approaches infinity
\cite[pp. 396--397]{muresan2008concrete}. Clearly, the latter grows
much faster than the former by an exponential factor of $6.93145^{N+1}$.
To manage the exploration of this space, we design a \texttt{\small split-and-merge}
Metropolis-Hastings procedure which iteratively visits all possible
ways of partitioning and ordering. The procedure randomly alternates
between the \texttt{\small split} move, where a subset is split into
two consecutive parts, and the \texttt{\small merge} move, where two
consecutive subsets are merged. The proposed model is termed Ordered
Sets Model ($\genModel$).

To discover hidden aspects in ordered sets (e.g., latent aspects that
capture the taste of a user in his or her movie genre), we further
introduce binary latent variables in a fashion similar to that of
restricted Boltzmann machines (RBMs) \cite{smolensky1986information}.
The posteriors of hidden units given the visible rank data can be
used as a vectorial representation of the data - this can be handy
in tasks such as computing distance measures or visualisation. This
results in a new model called Latent $\genModel$.

Finally, we show how the proposed Latent $\genModel$ can be applied
for collaborative filtering, e.g., when we need to take seen grouped
item ratings as input and produce a ranked list of unseen item for
each user. We then demonstrate and evaluate our model on large-scale
public datasets. The experiments show that our approach is competitive
against several state-of-the-art methods. 

The rest of the paper is organised as follows. Section~\ref{sec:Ordered-Set-Log-linear}
presents the log-linear model over ordered sets ($\genModel$) together
with our main contribution -- the \texttt{\small split-and-merge}
procedure. Section~\ref{sec:Introducing-Latent-Variables} introduces
Latent $\genModel$, which extends the $\genModel$ to incorporate
latent variables in the form of a set of binary factors. An application
of the proposed Latent $\genModel$ for collaborative filtering is
described in Section~\ref{sec:Application-in-Collaborative}. Related
work is reviewed in the next section, followed by the conclusions.

\section{Ordered Set Log-linear Models\label{sec:Ordered-Set-Log-linear}}

\begin{figure}
\begin{centering}
\begin{tabular}{ccc}
\includegraphics[width=0.3\textwidth]{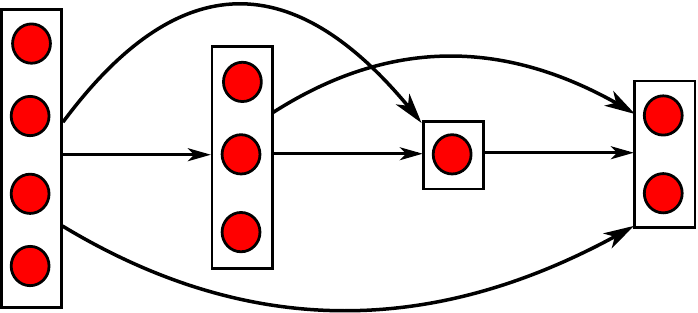}\qquad{} & \includegraphics[width=0.3\textwidth]{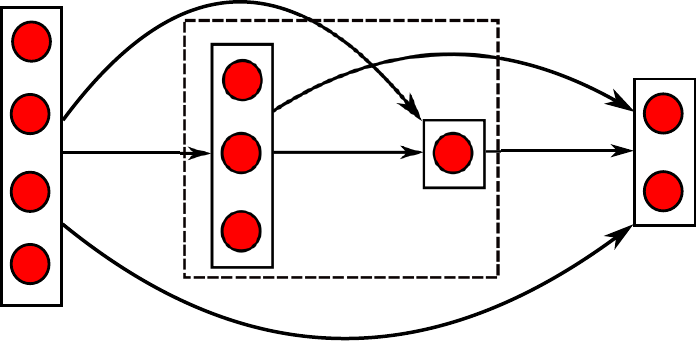} & \qquad{}\includegraphics[width=0.21\textwidth]{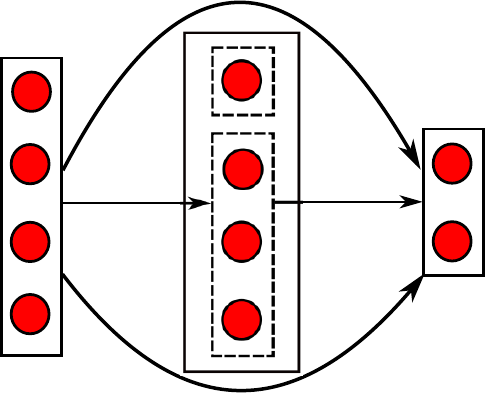}\tabularnewline
(a) OSM & (b)  & (c) \tabularnewline
\end{tabular}
\par\end{centering}

\caption{(a) Ordered Set Model; (b) the \texttt{\small split} operator; and
(c) the \texttt{\small merge} operator. The figure in (c) represents
the result of a merge of the middle two subsets in (b). Conversely,
the (b) figure can be considered as result of a splitting the middle
subset of the (c) figure. Arrows represent the preference orders,
not the causality or conditioning.\label{fig:Split-and-Merge}}
\end{figure}

\subsection{General Description}

We first present an intuitive description of the problem and our solutions
in modelling, learning and inference. Fig.~\ref{fig:Split-and-Merge}(a)
depicts the problem of grouping items into subsets (represented by
a box of circles) and ordering these subsets (represented by arrows
which indicate the ordering directions). This looks like a high-order
Markov chain in a standard setting, and thus it is tempting to impose
a chain-based distribution. However, the difficulty is that the partitioning
of set into subsets is also random, and thus a simple treatment is
not applicable. Recently, Truyen et al (2011) describe a model in
this direction with a careful treatment of the partitioning effect.
However, their model does not allow fast inference since we need to
take care of the high-order properties.

Our solution is as follows. To capture the grouping and relative ordering,
we impose on each group a subset potential function capturing the
relations among compatible elements, and on each pair of subsets a
ordering potential function. The distribution over the space of grouping
and ordering is defined using a log-linear model, where the product
of all potentials accounts for the unnormalised probability. This
log-linear parameterization allows flexible inference in the combinatorial
space of all possible groupings and orderings.

In this paper inference is carried out in a MCMC manner. At each step,
we randomly choose a \texttt{split} or a \texttt{merge}
operator. The \texttt{split} operator takes a subset
at random (e.g., Fig.~\ref{fig:Split-and-Merge}(c)) and uniformly
splits it into two smaller subsets. The order between these two smaller
subset is also random, but their relative positions with respect to
other subsets remain unchanged (e.g., Fig.~\ref{fig:Split-and-Merge}(b)).
The \texttt{merge} operator is the reverse (e.g.,
converting Fig.~\ref{fig:Split-and-Merge}(b) into Fig.~\ref{fig:Split-and-Merge}(c)).
With an appropriate acceptance probability, this procedure is guaranteed
to explore the entire combinatorial space.

Armed with this sampling procedure, learning can be carried out using
stochastic gradient techniques \cite{younes1989parametric}.

\subsection{Problem Description\label{sub:description}}

Given two objects $x_{i}$ and $x_{j}$, we use the notation $x_{i}\succ x_{j}$
to denote the expression of $x_{i}$ is ranked higher than $x_{j}$,
and $x_{i}\sim x_{j}$ to denote the between the two belongs to the
same group. Furthermore, we use the notation of $\objects=\left\{ x_{1},x_{2},\ldots,x_{N}\right\} $
as a collection of $N$ objects. Assume that $X$ is partitioned into
$T$ subsets $\left\{ X_{t}\right\} _{t=1}^{T}$. However, unlike
usual notion of partitioning of a set, we further posit an \emph{order}
among these subsets in which members of each subset presumably share
the same rank. Therefore, our partitioning process is \emph{order-sensitive}
instead of being exchangeable at the partition level. Specifically,
we use the indices $1,2,...,T$ to denote the \emph{decreasing} order
in the rank of subsets. These notations allow us to write the collection
of objects $\objects=\left\{ x_{1},\ldots,x_{N}\right\} $ as a union
of $T$ ordered subsets:%
\footnote{Alternatively, we could have proceeded from the permutation perspective
to indicate the ordering of the subsets, but we simplify the notation
here for clarity.%
} 
\begin{eqnarray}
\objects= & X_{1}\bigcup X_{2}\ldots\bigcup X_{T}\label{eq:partitions}
\end{eqnarray}
where $\{\objects_{t}\}_{t=1}^{T}$ are non-empty subsets of objects
so that $x_{i}\sim x_{j},\forall x_{i},x_{j}\in X_{t}$ $i\ne j,\forall t$.

As a special case when $T=N$, we obtain an exhaustive ordering among
objects wherein each subset has exactly one element and there is no
grouping among objects. This special case is equivalent with a complete
ranking scenario. To illustrate the complexity of the problem, let
us characterise the state-space, or more precisely, the number of
all possible ways of partitioning and ordering governed by the above
definition. Recall that there are $s\left(N,T\right)$ ways to divide
a set of $N$ objects into $T$ partitions, where $s\left(N,T\right)$
denotes the \emph{Stirling numbers of second kind} \cite[p. 105]{van1992course}.
Therefore, for each pair $(N,T)$, there are $s\left(N,T\right)T!$
ways to perform the partitioning with ordering. Considering all the
possible values of $T$ give us the size of our model state-space:
\begin{eqnarray}
\sum_{T=1}^{N}s\left(N,T\right)T! & =\mbox{Fubini}\left(N\right)=\sum_{k=1}^{\infty}\frac{k^{N}}{2^{k+1}}\label{eq:Fubini_number}
\end{eqnarray}
which is also known in combinatorics as the Fubini's number \cite[pp. 396--397]{muresan2008concrete}.
This number grows super-exponentially and it is known that it approaches
$N!/(2\left(\log2\right)^{N+1})$ as $N\rightarrow\infty$ \cite[pp. 396--397]{muresan2008concrete}.
Taking the logarithm, we get $\log N!-(N+1)\log\log2-\log2$. As $\log\log2<0$,
this clearly grows faster than $\log N!$, which is the log of the
size of the standard complete permutations.

\subsection{Model Specification}

Denote by $\Phi(X_{t})\in\Real^{+}$ a positive potential function
over a single subset%
\footnote{In this paper, we do not consider the case of empty sets, but it can
be assumed that $\phi(\emptyset)=1$.%
} $X_{t}$ and by $\Psi(X_{t}\succ X_{t'})\in\Real^{+}$a potential
function over a ordered pair of subsets $\left(X_{t},X_{t'}\right)$
where $t<t'$. Our intention is to use $\Phi(X_{t})$ to encode the
compatibility among all member of $X_{t}$, and $\Psi(X_{t}\succ X_{t'})$
to encode the ordering properties between $X_{t}$ and $X_{t'}$.
We then impose a distribution over the collection of objects as:

\begin{eqnarray}
P(X) & = & \frac{1}{Z}\Omega(X),\mbox{\qquad where}\,\,\Omega(X)=\prod_{t}\Phi(X_{t})\prod_{t'>t}\Psi(X_{t}\succ X_{t'})\label{eq:model-def}
\end{eqnarray}
and $Z=\sum_{X}\Omega(X)$ is the partition function. We further posit
the following factorisation for the potential functions:

\begin{eqnarray}
\Phi(X_{t}) & = & \prod_{i,j\in X_{t}|j>i}\varphi(x_{i}\sim x_{j});\mbox{\qquad\qquad}\Psi(X_{t}\succ X_{t'})=\prod_{i\in X_{t}}\prod_{j\in X_{t'}}\psi(x_{i}\succ x_{j})\label{eq:potent-factorise}
\end{eqnarray}
where $\varphi(x_{i}\sim x_{j})\in\Real^{+}$ captures the effect
of \emph{grouping}, and $\psi(x_{i}\succ x_{j})\in\Real^{+}$ captures
the relative ordering between objects $x_{i}$ and $x_{j}$. Hereafter,
we shall refer to this proposed model as the \emph{Ordered Set Model}
(OSM).

\subsection{\texttt{\small Split-and-Merge} MCMC Inference}

In order to evaluate $P(X)$ we need to sum over all possible configurations
of $X$ which is in the complexity of the Fubini($N$) over the set
of $N$ objects (cf. Section \ref{sub:description}, Eq. \ref{eq:Fubini_number}).
We develop a Metropolis-Hastings (MH) procedure for sampling $P(X)$.
Recall that the MH sampling involves a proposal distribution $Q$
that allows drawing a new sample $X'$ from the current state $X$
with probability $Q(X'|X)$. The move is then accepted with probability
\begin{equation}
P_{\mathsf{accept}}=\min\left\{ 1,l\times p\right\} ,\mbox{\qquad where}\,\, l=\frac{P(X')}{P(X)}=\frac{\Omega(X')}{\Omega(X)}\,\,\mbox{and}\,\, p=\frac{Q(X|X')}{Q(X'|X)}\label{eq:acceptance}
\end{equation}
To evaluate the likelihood ratio $l$ we use the model specification
defined in Eq (\ref{eq:model-def}). We then need to compute the proposal
probability ratio $p$. The key intuition is to design a random local
move from $X$ to $X'$ that makes a relatively small change to the
current partitioning and ordering. If the change is large, then the
rejection rate is high, thus leading to high cost (typically the computational
cost increases with the step size of the local moves). On the other
hand, if the change is small, then the random walks will explore the
state-space too slowly.

We propose two operators to enable the proposal move: the \texttt{\small split}
operator takes a non-singleton subset $X_{t}$ and randomly splits
it into two sub-subsets $\{X_{t}^{1},X_{t}^{2}\}$, where $X_{t}^{2}$
is inserted \emph{right} next to $X_{t}^{1}$; and the \texttt{\small merge}
operator takes two \emph{consecutive} subsets $\{X_{t},X_{t+1}\}$
and merges them. This dual procedure will guarantee exploration of
all possible configurations of partitioning and ordering, given enough
time (See Figure~\ref{fig:Split-and-Merge} for an illustration).

\subsubsection{\texttt{\small Split}\emph{ }Operator}

Assume that among the $T$ subsets, there are $T_{\mathsf{split}}$
non-singleton subsets from which we randomly select one subset to
split, and let this be $X_{t}$. Since we want the resulting sub-subsets
to be non-empty, we first randomly draw two distinct objects from
$X_{t}$ and place them into the two subsets. Then, for each remaining
object, there is an equal chance going to either $X_{t}^{1}$ or $X_{t}^{2}$.
Let $N_{t}=|X_{t}|$, the probability of this drawing is $\left(N_{t}(N_{t}-1)2^{N_{t}-2}\right)^{-1}$.
Since the probability that these two sub-subsets will be merged back
is $T^{-1}$, the proposal probability ratio $p_{\mathsf{split}}$
can be computed as in Eq (\ref{eq:ratio-split}). Since our potential
functions depend only on the relative orders between subsets and between
objects in the same set, the likelihood ratio $l_{\mathsf{split}}$
due to the \texttt{\small split} operator does not depend on other
subsets, it can be given as in Eq (\ref{eq:energy-change-split}).
This is because the members of $X_{t}^{1}$ are now ranked higher
than those of $X_{t}^{2}$ while they are of the same rank previously. 

\begin{center}
\begin{minipage}[t]{0.5\columnwidth}%
\begin{equation}
p_{\mathsf{split}}=\frac{T_{\mathsf{split}}N_{t}(N_{t}-1)2^{N_{t}-2}}{T}\label{eq:ratio-split}
\end{equation}
\end{minipage}%
\begin{minipage}[t]{0.5\columnwidth}%
\begin{equation}
l_{\mathsf{split}}=\prod_{x_{i}\in X_{t}^{1}}\prod_{x_{j}\in X_{t}^{2}}\frac{\psi(x_{i}\succ x_{j})}{\varphi(x_{i}\sim x_{j})}\label{eq:energy-change-split}
\end{equation}
\end{minipage}
\par\end{center}

\subsubsection{\texttt{\small Merge}\emph{ }Operator}

For $T$ subsets, the probability of merging two consecutive ones
will be $(T-1)^{-1}$ since there are $T-1$ pairs, and each pair
can be merged in exactly one way. Let $T_{\mathsf{merge}}$ be the
number of non-singleton subsets after the merge, and let $N_{t}$
and $N_{t+1}$be the sizes of the two subsets $X_{t}$ and $X_{t+1}$,
respectively. Let $N_{t}^{*}=N_{t}+N_{t+1}$, the probability of recovering
the state before the merge (by applying the \texttt{\small split}
operator) is $\left(T_{\mathsf{merge}}N_{t}^{*}(N_{t}^{*}-1)2^{N_{t}^{*}-2}\right)^{-1}$.
Consequently, the proposal probability ratio $p_{\mathsf{merge}}$
can be given as in Eq (\ref{eq:ratio-merge}), and the likelihood
ratio $l_{\mathsf{merge}}$ is clearly the inverse of the split\emph{
}case as shown in Eq (\ref{eq:energy-change-merge}).

\begin{center}
\begin{minipage}[t]{0.5\columnwidth}%
\begin{equation}
p_{\mathsf{merge}}=\frac{T-1}{T_{\mathsf{merge}}N_{t}^{*}(N_{t}^{*}-1)2^{N_{t}^{*}-2}}\label{eq:ratio-merge}
\end{equation}
\end{minipage}%
\begin{minipage}[t]{0.5\columnwidth}%
\begin{equation}
l_{\mathsf{merge}}=\prod_{x_{i}\in X_{t}}\prod_{x_{j}\in X_{t+1}}\frac{\varphi(x_{i}\sim x_{j})}{\psi(x_{i}\succ x_{j})}\label{eq:energy-change-merge}
\end{equation}
\end{minipage}
\par\end{center}

Finally, the pseudo-code of the \texttt{\small split-and-merge} Metropolis-Hastings
procedure for the $\genModel$ is presented in Algorithm~\ref{alg:split-merge-MH}. 

\begin{algorithm}
1. Given an initial state $X$.

2. \textbf{Repeat} until convergence

~%
\begin{tabular}{|>{\raggedright}p{0.03\columnwidth}|>{\raggedright}p{0.8\columnwidth}}
\multicolumn{2}{|l}{\quad{}2a. Draw a random number $\eta\in[0,1]$.}\tabularnewline
\multicolumn{2}{|l}{\quad{}2b. \textbf{If} $\eta<0.5$ \{\texttt{\small Split}\}}\tabularnewline
 & i. Randomly choose a non-singleton subset.

ii. Split into two sub-subsets and insert one sub-subset right after
the another.

iii. Evaluate the acceptance probability $P_{\mathsf{accept}}$ using
Eqs.(\ref{eq:ratio-split},\ref{eq:energy-change-split},\ref{eq:acceptance}).

iv. Accept the move with probability $P_{\mathsf{accept}}$.\tabularnewline
\multicolumn{2}{|l}{\quad{}\textbf{Else} \{\texttt{\small Merge}\}}\tabularnewline
 & i. Randomly choose two consecutive subsets.

ii. Merge them in one, keeping the relative orders with other subsets
unchanged.

iii. Evaluate the acceptance probability $P_{\mathsf{accept}}$ using
Eqs.(\ref{eq:ratio-merge},\ref{eq:energy-change-merge},\ref{eq:acceptance}).

iv. Accept the move with probability $P_{\mathsf{accept}}$.\tabularnewline
\multicolumn{2}{|l}{\quad{}\textbf{End}}\tabularnewline
\multicolumn{2}{|c}{}\tabularnewline
\end{tabular}

\textbf{End}

\caption{Pseudo-code of the \texttt{\small split-and-merge} Metropolis-Hastings
for $\genModel$.\label{alg:split-merge-MH}}
\end{algorithm}

\subsection{Estimating Partition Function \label{sub:Partition-Function-Learning-Predict}}

To estimate the normalisation constant $Z$, we employ an efficient
procedure called Annealed Importance Sampling (AIS) proposed recently
\cite{neal2001annealed}. More specifically, AIS introduces the notion
of inverse-temperature $\tau$ into the model, that is $P(X|\tau)\propto\Omega(X)^{\tau}$.

Let $\{\tau_{s}\}_{s=0}^{S}$ be the (slowly) increasing sequence
of temperature, where $\tau_{0}=0$ and $\tau_{S}=1$, that is $\tau_{0}<\tau_{1}...<\tau_{S}$.
At $\tau_{0}=0$, we have a uniform distribution, and at $\tau_{S}=1$,
we obtain the desired distribution. At each step $s$, we draw a sample
$X^{s}$ from the distribution $P(X|\tau_{s-1})$ (e.g. using the
\texttt{\small split-and-merge} procedure). Let $P^{*}(X|\tau)$ be
the unnormalised distribution of $P(X|\tau)$, that is $P(X|\tau)=P^{*}(X|\tau)/Z(\tau)$.
The final weight after the annealing process is computed as
\begin{equation}
w =\frac{P^{*}(X^{1}|\tau_{1})}{P^{*}(X^{1}|\tau_{0})}\frac{P^{*}(X^{2}|\tau_{2})}{P^{*}(X^{2}|\tau_{1})}...\frac{P^{*}(X^{S}|\tau_{S})}{P^{*}(X^{S}|\tau_{S-1})}
\end{equation}

The above procedure is repeated $R$ times. Finally, the normalisation
constant at $\tau=1$ is computed as $Z(1)\approx Z(0)\left(\sum_{r=1}^{R}w^{(r)}/R\right)$where
$Z(0)=\mbox{Fubini}(N)$, which is the number of configurations of
the model state variables $X$.

\subsection{Log-linear Parameterisation and Learning \label{sub:Log-linear-Parameterisation}}

Here we assume that the model is in the log-linear form, that is $\varphi(x_{i}\sim x_{j})=\exp\left\{ \sum_{a}\alpha_{a}f_{a}(x_{i},x_{j})\right\} $
and $\psi(x_{i}\succ x_{j})=\exp\left\{ \sum_{b}\beta_{b}g_{b}(x_{i},x_{j})\right\} $,
where $\left\{ f_{a}(\cdot),\, g_{b}(\cdot)\right\} $ are sufficient
statistics (or feature functions) and $\left\{ \alpha_{a},\,\beta_{b}\right\} $
are free parameters.

Learning by maximising (log-)likelihood in log-linear models with
respect to free parameters often leads to computing the expectation
of sufficient statistics. For example, $\left\langle f_{a}(x_{i},x_{j})\right\rangle _{P(x_{i}\sim x_{j})}$
is needed in the gradient of the log-likelihood with respect to $\alpha_{a}$,
where $P(x_{i}\sim x_{j})$ is the pairwise marginal. Unfortunately,
computing $P(x_{i}\sim x_{j})$ is inherently hard, and running a
full MCMC chain to estimate it is too expensive for practical purposes.
Here we follow the stochastic approximation proposed in \cite{younes1989parametric},
in that we iteratively update parameters after very short MCMC chains
(e.g., using Algorithm~\ref{alg:split-merge-MH}).

\section{Introducing Latent Variables to OSMs \label{sec:Introducing-Latent-Variables}}

\begin{figure*}
\begin{centering}
\begin{tabular}{ccc}
\includegraphics[width=0.4\textwidth]{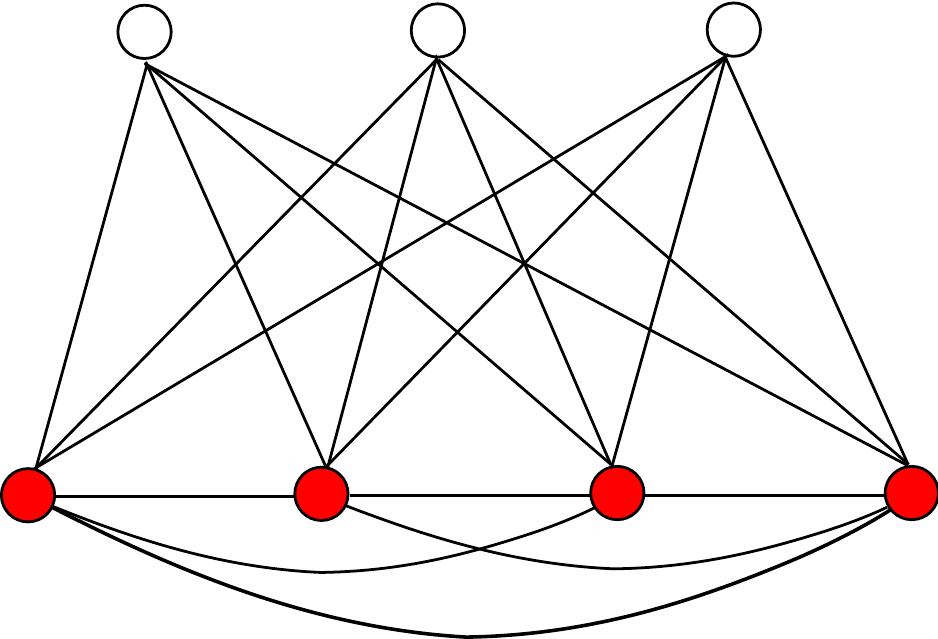} & \qquad{}\qquad{} & \includegraphics[width=0.35\textwidth]{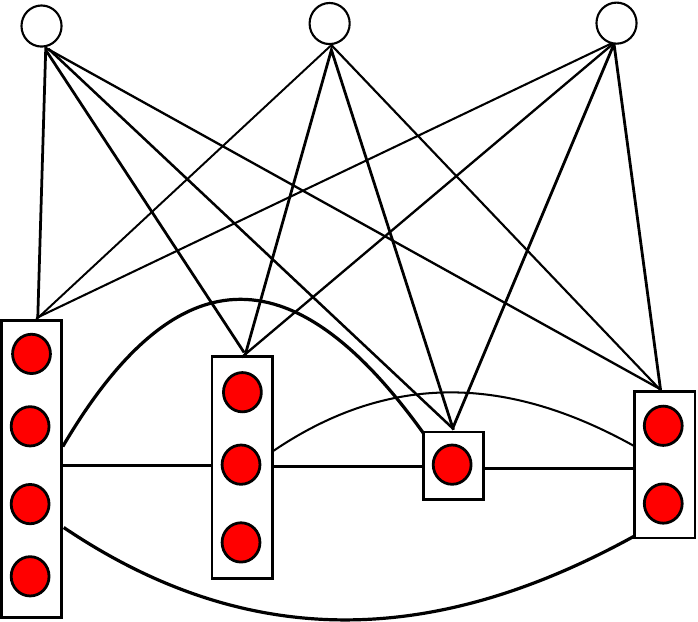}\tabularnewline
(a) &  & (b)\tabularnewline
\end{tabular}
\par\end{centering}

\caption{(a) A Semi-Restricted Boltzmann Machine representation of vectorial
data: each shaded node represents a visible variable and empty nodes
the hidden units. (b) A Latent $\genModel$ for representing ordered
sets: each box represents a subset of objects. \label{fig:models}}
\end{figure*}

In this section, we further extend the proposed $\genModel$ by introducing
latent variables into the model. The latent variables serve multiple
purposes. For example, in collaborative filtering, each person chooses
only a small subset of objects, thus the specific choice of objects
and the ranking reflects personal taste. This cannot be discovered
by the standard $\genModel$. Second, if we want to measure the distance
or similarity between two ordered partitioned sets, e.g. for clustering
or visualisation, it may be useful to first transform the data into
some vectorial representation.

\subsection{Model Specification}

Denote by $\hidden=(h_{1},h_{2},...,h_{K})\in\{0,1\}^{K}$ the hidden
units to be used in conjunction with the ordered sets. The idea is
to estimate the posterior $P(h_{k}=1|X)$ - the probability that the
$k^{\mbox{th}}$ hidden unit will be activated by the input $X$.
Thus, the requirement is that the model should allow the evaluation
of $P(h_{k}=1\mid X)$ efficiently. Borrowing from the Restricted
Boltzmann Machine architecture \cite{smolensky1986information,welling2005efh},
we can extend the model potential function as follows:
\begin{equation}
\hat{\Omega}(X,\hidden)=\Omega(X)\prod_{k}\Omega_{k}(X)^{h_{k}}\label{eq:latent-joint-potential}
\end{equation}
where $\Omega_{k}(X)$ admits the similar factorisation as $\Omega(X)$,
i.e. $\Omega_{k}(X)=\prod_{t}\Phi_{k}(X_{t})\prod_{t'>t}\Psi_{k}(X_{t}\succ X_{t'})$,
and

\begin{eqnarray}
\Phi_{k}(X_{t}) & = & \prod_{i,j\in X_{t}|j>i}\varphi_{k}(x_{i}\sim x_{j});\quad\quad\Psi_{k}(X_{t}\succ X_{t'})=\prod_{i\in X_{t}}\prod_{j\in X_{t'}}\psi_{k}(x_{i}\succ x_{j})\label{eq:hidden-potent-factorise}
\end{eqnarray}
where $\varphi_{k}(x_{i}\sim x_{j})$ and $\psi_{k}(x_{i}\succ x_{j})$
capture the events of tie and relative ordering between objects $x_{i}$
and $x_{j}$ \emph{under the presence of the $k^{\mbox{th}}$ hidden
unit}, respectively.

We then define the model with hidden variables as $P(X,\hidden)=\hat{\Omega}(X,\hidden)/Z$,
where $Z=\sum_{X,\hidden}\hat{\Omega}(X,\hidden)$. A graphical representation
is given in Figure~\ref{fig:models}b. Hereafter, we shall refer
to this proposed model as the Latent $\genModel$.

\subsection{Inference}

The posteriors are indeed efficient to evaluate:

\begin{eqnarray}
P(\hidden\mid X) & = & \prod_{k}P(h_{k}\mid X),\mbox{\quad where\,\,}P(h_{k}=1\mid X)=\frac{1}{1+\Omega_{k}(X)^{-1}}\label{eq:hidden-cond-factorise}
\end{eqnarray}
Denote by $h_{k}^{1}$ as the shorthand for $h_{k}=1$, the vector
$(P(h_{1}^{1}\mid X),P(h_{2}^{1}\mid X),...,P(h_{K}^{1}\mid X))$
can then be used as a latent representation of the configuration $X$.

The generation of $X$ given $\hidden$ is, however, much more involved
as we need to explore the whole subset partitioning and ordering space:
\begin{equation}
P(X\mid\hidden)=\frac{\hat{\Omega}(X,\hidden)}{\sum_{X}\hat{\Omega}(X,\hidden)}=\frac{\Omega(X)\prod_{k}\Omega_{k}(X)^{h_{k}}}{\sum_{X}\Omega(X)\prod_{k}\Omega_{k}(X)^{h_{k}}}\label{eq:visible-cond}
\end{equation}

For inference, since we have two layers $X$ and $\hidden$, we can
alternate between them in a Gibbs sampling manner, that is, sampling
$X$ from $P(X\mid\hidden)$ and then $\hidden$ from $P(\hidden\mid X)$.
Since sampling from $P(\hidden\mid X)$ is straightforward, it remains
to sample from $P(X|\hidden)=\hat{\Omega}(X,\hidden)/\sum_{X}\hat{\Omega}(X,\hidden)$.
Since $\hat{\Omega}(X,\hidden)$ has the same factorisation structure
into a product of pairwise potentials as $\Omega(X)$, we can employ
the \texttt{\small split-and-merge} technique described in the previous
section in a similar manner.

To see how, let $\hat{\varphi}(x_{i}\sim x_{j},\hidden)=\varphi(x_{i}\sim x_{j})\prod_{k}\varphi_{k}(x_{i}\sim x_{j})^{h_{k}}$
and $\hat{\psi}(x_{i}\succ x_{j},\hidden)=\psi_{k}(x_{i}\succ x_{j})\prod_{k}\psi_{k}(x_{i}\succ x_{j})^{h_{k}}$,
then from Eqs.(\ref{eq:potent-factorise},\ref{eq:latent-joint-potential},\ref{eq:hidden-potent-factorise}).
We can see that $\hat{\Omega}(X,\hidden)$ is now factorised into
products of $\hat{\varphi}(x_{i}\sim x_{j},\hidden)$ and $\hat{\psi}(x_{i}\succ x_{j},\hidden)$
in the same way as $\Omega(X)$ into products of $\varphi(x_{i}\sim x_{j})$
and $\psi(x_{i}\succ x_{j})$:

\begin{eqnarray*}
\hat{\Omega}(X,\hidden) & = & \Omega(X)\prod_{k}\Omega_{k}(X)^{h_{k}}=\prod_{t}\hat{\Phi}(X_{t},\hidden)\prod_{t'>t}\hat{\Psi}(X_{t}\succ X_{t'},\hidden)
\end{eqnarray*}
where
\begin{eqnarray*}
\hat{\Phi}(X_{t},\hidden) & = & \prod_{i,j\in X_{t}|j>i}\hat{\varphi}(x_{i}\sim x_{j},\hidden);\quad\quad\hat{\Psi}(X_{t}\succ X_{t'},\hidden)=\prod_{i\in X_{t}}\prod_{j\in X_{t'}}\hat{\psi}(x_{i}\succ x_{j},\hidden)
\end{eqnarray*}

Estimating the normalisation constant $Z$ can be performed using
the AIS procedure described earlier (cf. Section \ref{sub:Partition-Function-Learning-Predict}),
except that the unnormalised distribution $P^{*}(X|\tau)$ is given
as: 
\begin{eqnarray*}
P^{*}(X\mid\tau) & =\sum_{\hidden}\hat{\Omega}(X,\hidden)^{\tau}= & \Omega(X)^{\tau}\prod_{k}\left(1+\Omega_{k}(X)^{\tau}\right)
\end{eqnarray*}
which can be computed efficiently for each $X$. 

For sampling $X^{s}$ from $P(X\mid\tau_{s-1})$, one way is to sample
directly from the $P(X\mid\tau_{s-1})$ in a Rao-Blackwellised fashion
(e.g. by marginalising over $\hidden$ we obtain the unnormalised
$P^{*}(X|\tau)$). A more straightforward way is alternating between
$X\mid\hidden$ and $\hidden\mid X$ as usual. Although the former
would give lower variance, we implement the latter for simplicity.
The remaining is similar to the case without hidden variables, and
we note that the base partition function $Z(0)$ should be modified
to $Z(0)=\mbox{Fubini}(N)2^{K}$, taking into account of $K$ binary
hidden variables. A pseudo-code for the \texttt{\small split-and-merge}
algorithm for Latent $\genModel$ is given in Algorithm~\ref{eq:ratio-merge}.

\begin{algorithm}
1. Given an initial state $X$.

2. \textbf{Repeat} until convergence

~%
\begin{tabular}{|>{\raggedright}p{0.03\columnwidth}||>{\raggedright}p{0.03\columnwidth}||>{\raggedright}p{0.8\columnwidth}}
\multicolumn{3}{|l}{\quad{}2a. Sample $\hidden$ from $P(\hidden\mid X)$ using Eq.(\ref{eq:hidden-cond-factorise}).}\tabularnewline
\multicolumn{3}{|l}{\quad{}2b. Sample $X$ from $P(X\mid\hidden)$ using Eq.(\ref{eq:visible-cond})
and Algorithm~\ref{alg:split-merge-MH}.}\tabularnewline
\multicolumn{3}{|l}{\quad{}\textbf{End}}\tabularnewline
\end{tabular}

\textbf{End}

\caption{Pseudo-code of the \texttt{\small split-and-merge} Gibbs/Metropolis-Hastings
for Latent $\genModel$.\label{alg:spit-merge-Gibbs/MH}}
\end{algorithm}

\subsection{Parameter Specification and Learning \label{sub:Latent-OSM-Parameter-Specification}}

Like the $\genModel$, we also assume log-linear parameterisation.
In addition to those potentials shared with the $\genModel$, here
we specify hidden-specific potentials as follows: $\varphi_{k}(x_{i}\sim x_{j})^{h_{k}}=\exp\left\{ \sum_{a}\lambda_{ak}f_{a}(x_{i},x_{j})h_{k}\right\} $
and $\psi_{k}(x_{i}\succ x_{j})^{h_{k}}=\exp\left\{ \sum_{b}\mu_{bk}g_{b}(x_{i},x_{j})h_{k}\right\} $.
Now $f_{a}(x_{i},x_{j})h_{k}$ and $g_{b}(x_{i},x_{j})h_{k}$
are new sufficient statistics. As before, we need to estimate the
expectation of sufficient statistics, e.g., $\left\langle f_{a}(x_{i},x_{j})h_{k}\right\rangle _{P(x_{i},x_{j},h_{k})}$.
Equipped with Algorithm~\ref{eq:ratio-merge}, the stochastic gradient
trick as in Section~\ref{sub:Log-linear-Parameterisation}  can then
be used, that is, parameters are updated after very short chains (with
respect to the model distribution $P(X,\hidden)$).

\section{Application in Collaborative Filtering\label{sec:Application-in-Collaborative} }

In this section, we present one specific application of our Latent
$\genModel$ in collaborative filtering. Recall that in this application,
each user has usually expressed their preferences over a set of items
by rating them (e.g., by assigning each item a small number of stars).
Since it is cumbersome to rank all the items completely, the user
often joins items into groups of similar ratings. As each user often
rates only a handful of items out of thousands (or even millions),
this creates a sparse ordering of subsets. Our goal is to first discover
the latent taste factors for each user from their given ordered subsets,
and then use these factors to recommend new items for each individual.

\subsection{Rank Reconstruction and Completion}

In this application, we are limited to producing a \emph{complete
ranking} over objects instead of subset partitioning and ordering.
Here we consider two tasks: (i) \emph{rank completion} where we want
to rank unseen items given a partially ranked set%
\footnote{This is important in recommendation, as we shall see in the experiments.%
}, and (ii) \emph{rank reconstruction}%
\footnote{This would be useful in data compression setting.%
} where we want to reconstruct the complete rank $\hat{X}$ from the
posterior vector $(P(h_{1}^{1}\mid X),P(h_{2}^{1}\mid X)..,,P(h_{K}^{1}\mid X))$.

\paragraph{Rank completion.}

Assume that an unseen item $x_{j}$ might be ranked higher than any
seen item $\{x_{i}\}_{i=1}^{N}$. Let us start from the mean-field
approximation
\begin{eqnarray*}
P(x_{j}\mid X) & = & \sum_{\hidden}P(x_{j},\hidden\mid X)\approx Q_{j}(x_{j}\mid X)\prod_{k}Q_{k}(h_{k}\mid X)
\end{eqnarray*}
From the mean-field theory, we arrive at Eq (\ref{eq:meanfield}),
which resembles the factorisation in (\ref{eq:latent-joint-potential}).
\begin{eqnarray}
Q_{j}(x_{j}\mid X) & \propto & \Omega(x_{j},X)\prod_{k}\Omega_{k}(x_{j},X)^{Q_{k}(h_{k}^{1}|X)}\label{eq:meanfield}
\end{eqnarray}
Now assume that $X$ is sufficiently informative to estimate $Q_{k}(h_{k}\mid X)$,
we make further approximation $Q_{k}(h_{k}^{1}\mid\state)\approx P(h_{k}^{1}\mid\state)$.
Finally, due to the factorisation in (\ref{eq:hidden-potent-factorise}),
this reduces to 
\[
Q_{j}(x_{j}\mid X)\propto\prod_{i}\left[\psi(x_{j}\succ x_{i})\prod_{k}\psi_{k}(x_{j}\succ x_{i})^{P_{k}(h_{k}^{1}\mid X)}\right]
\]
The RHS can be used for the purpose of ranking among new items $\{x_{j}\}$.

\paragraph{Rank reconstruction.}

The \emph{rank reconstruction} task can be thought as estimating $\hat{X}=\arg\max_{X'}Q(X'|X)$
where $Q(X'|X)=\sum_{\hidden}P(X'|\hidden)P(\hidden|X)$. Since this
maximisation is generally intractable, we may approximate it by treating
$X'$ as state variable of \emph{unseen} items, and apply the mean-field
technique as in the completion task.

\subsection{Models Implementation}

To enable fast recommendation, we use a rather simple scheme: Each
item is assigned a worth $\phi(x_{i})\in\Real^{+}$ which can be used
for ranking purposes. Under the Latent OSM, the worth is also associated
with a hidden unit, e.g. $\phi_{k}(x_{i})$. Then the events of grouping
and ordering can be simplified as
\begin{eqnarray*}
\varphi_{k}(x_{i}\sim x_{j}) & = & \theta\sqrt{\phi_{k}(x_{i})\phi_{k}(x_{j})};\mbox{\quad and\quad}\psi_{k}(x_{i}\succ x_{j})=\phi_{k}(x_{i})
\end{eqnarray*}
where $\theta>0$ is a factor signifying the contribution of item
compatibility to the model probability. Basically the first equation
says that if the two items are compatible, their worth should be positively
correlated. The second asserts that if there is an ordering, we should
choose the better one. This reduces to the tie model of \cite{davidson1970extending}
when there are only two items.

For learning, we parameterise the models as follows
\begin{eqnarray*}
\theta=e^{\nu};\mbox{\quad\quad}\phi(x_{i}) & = & e^{u_{i}};\mbox{\quad\quad}\phi_{k}(x_{i})=e^{W_{ik}}
\end{eqnarray*}
where $\nu$, $\{u_{i}\}$ and $\{W_{ik}\}$ are free parameters.
The Latent $\genModel$ is trained using stochastic gradient with
a few samples per user to approximate the gradient (e.g., see Section~\ref{sub:Latent-OSM-Parameter-Specification}).
To speed up learning, parameters are updated after every block of
$100$ users. Figure~\ref{fig:MovieLens-results}(a) shows the learning
progress with learning rate of $0.01$ using parallel \emph{persistent}
Markov chains, one chain per user \cite{younes1989parametric}.
The samples get closer to the observed data as the model is updated,
while the acceptance rates of the \texttt{\small split-and-merge}
decrease, possibly because the samplers are near the region of attraction.
A notable effect is that the \texttt{\small split-and-merge} dual
operators favour sets of small size due to the fact that there are
far more many ways to split a big subset than to merge them. For the
AIS, we follow previous practice (e.g. see \cite{salakhutdinov2008quantitative}),
i.e. $S=\{10^{3},10^{4}\}$ and $R=\{10,100\}$.

For comparison, we implemented existing methods including the Probabilistic
Matrix Factorisation (\emph{PMF}) \cite{salakhutdinov2008probabilistic}
where the predicted rating is used as scoring function, the Probabilistic
Latent Preference Analysis (\emph{pLPA}) \cite{liu2009probabilistic},
the ListRank.MF \cite{shi2010list} and the matrix-factored Plackett-Luce
model \cite{Truyen:2011a} (\emph{Plackett-Luce.MF}). For the pLPA
we did not use the MM algorithm but resorted to simple gradient ascent
for the inner loop of the EM algorithm. We also ran the CoFi$^{RANK}$
variants \cite{weimer2008cofi} with code provided by the authors%
\footnote{http://cofirank.org%
}. We found that the ListRank-MF and the Plackett-Luce.MF are very
sensitive to initialisation, and good results can be obtained by randomly
initialising the user-based parameter matrix with non-negative entries.
To create a rank for Plackett-Luce.MF, we order the ratings according
to \texttt{\small quicksort}.

The performance will be judged based on the correlation between the
predicted rank and ground-truth ratings. Two performance metrics are
reported: the Normalised Discounted Cumulative Gain at the truncated
position $T$ (NDCG$@T$) \cite{jarvelin2002cumulated}, and the Expected
Reciprocal Rank (ERR) \cite{chapelle2009expected}: 

\texttt{\small 
\begin{eqnarray*}
\mbox{NDCG}@T & =\frac{1}{\kappa(T)} & \sum_{i=1}^{T}\frac{2^{r_{i}}-1}{\log_{2}(1+i)};\mbox{\quad}\mbox{ERR}=\sum_{i}\frac{1}{i}V(r_{i})\prod_{j=1}^{i-1}(1-V(r_{j}))\mbox{\quad for\,\,}V(r)=\frac{2^{r-1}-1}{16}
\end{eqnarray*}
}where $r_{i}$ is the relevance judgment of the movie at position
$i$, $\kappa(T)$ is a normalisation constant to make sure that the
gain is $1$ if the rank is correct. Both the metrics put more emphasis
on top ranked items.

\subsection{Results}

\begin{figure*}
\begin{centering}
\begin{tabular}{ccc}
\includegraphics[width=0.3\textwidth,height=0.3\textwidth]{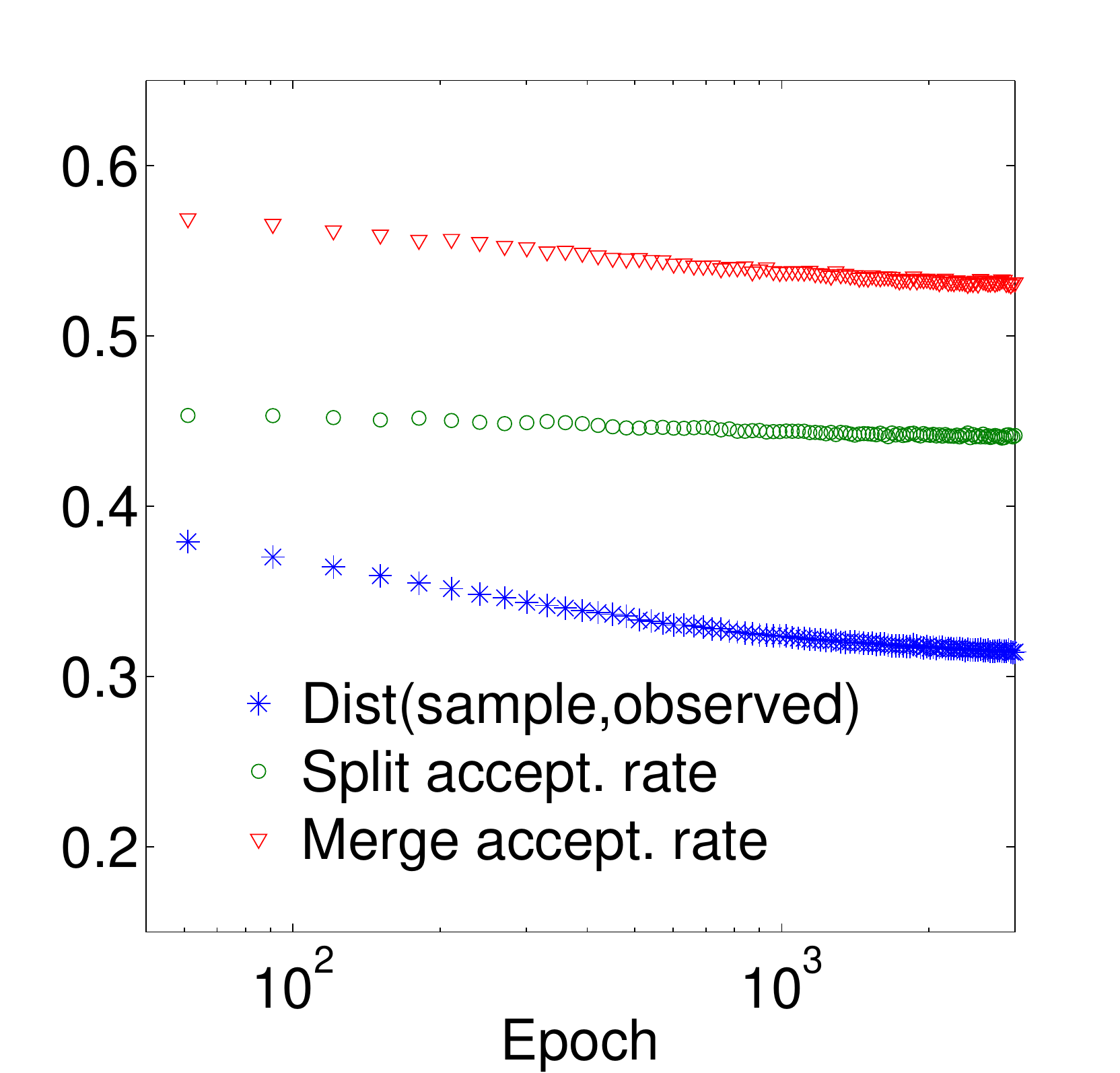} & \includegraphics[width=0.3\textwidth,height=0.3\textwidth]{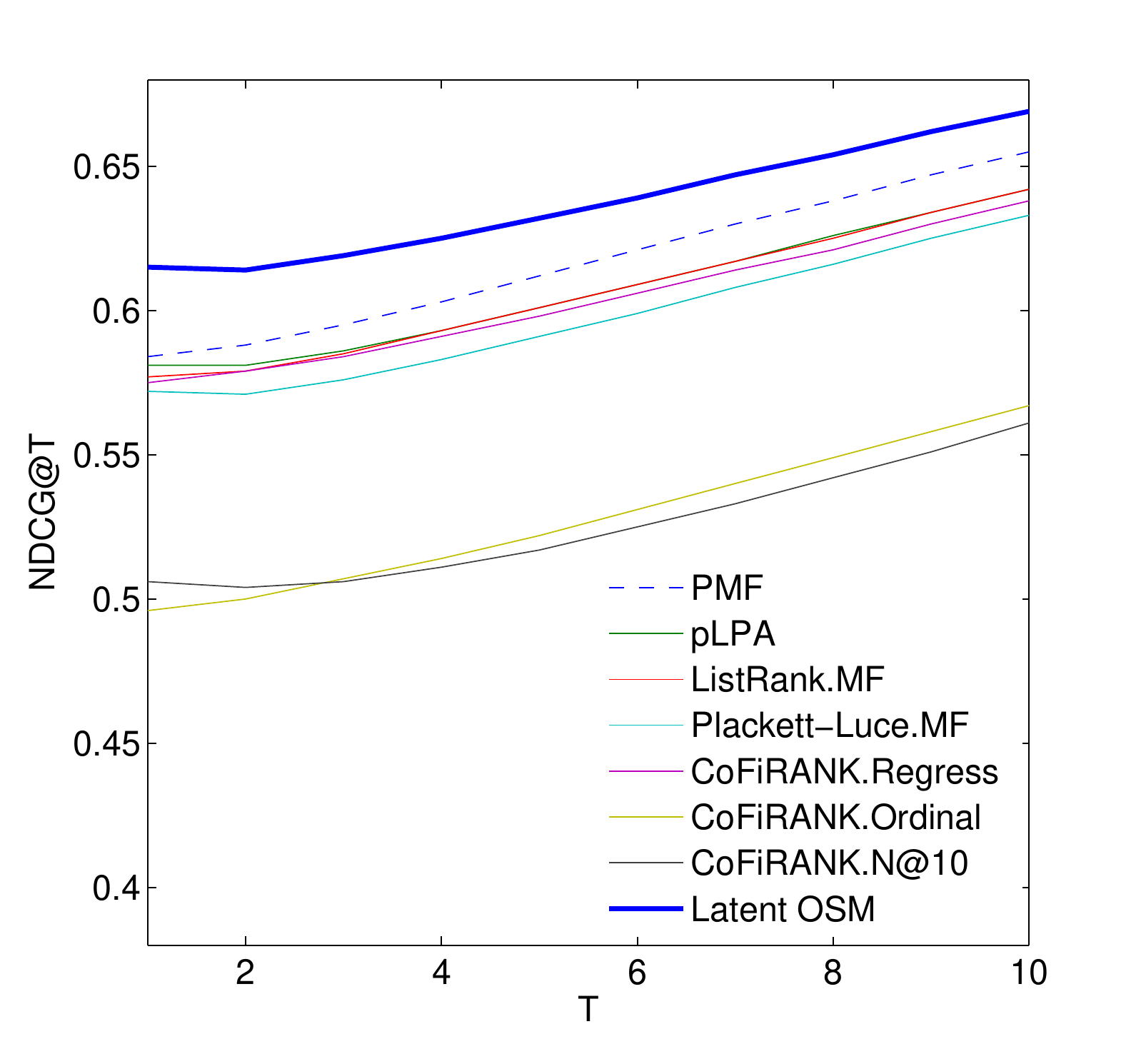} & \includegraphics[width=0.3\textwidth,height=0.3\textwidth]{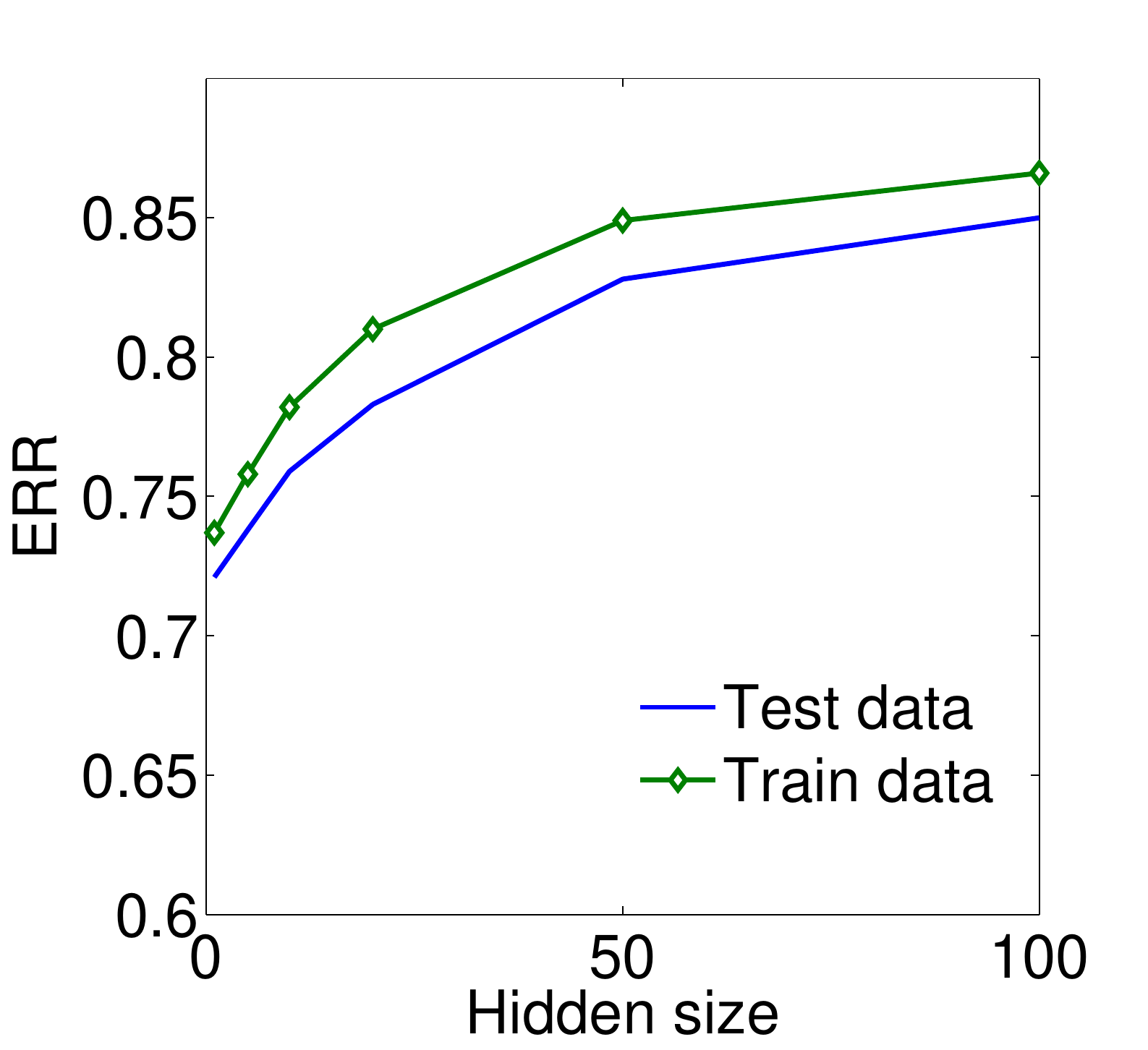}\tabularnewline
{\small (a)} & {\small (b)} & {\small (c)}\tabularnewline
\end{tabular}
\par\end{centering}

\caption{Results with MovieLens data. (a) Learning progress with time: Dist(sample,observed)
is the portion of pairwise orders being incorrectly sampled by the
\texttt{\small split-and-merge} Markov chains ($N=10$,$K=20$). (b)
Rank completion, as measured in NDCG@T ($N=20,\, K=50$). (c) Rank
reconstruction ($N=10$) - trained on $9,000$ users and tested on
$1,000$ users. \label{fig:MovieLens-results}}
\end{figure*}

We evaluate our proposed model and inference on large-scale collaborative
filtering datasets: the MovieLens%
\footnote{http://www.grouplens.org/node/12%
} $10$M and the Netflix challenge%
\footnote{http://www.netflixprize.com%
}. The MovieLens dataset consists of slightly over $10$ million half-integer
ratings (from $0$ to $5$) applied to $10,681$ movies by $71,567$
users. The ratings are from $0.5$ to $5$ with $0.5$ increments.
We divide the rating range into $5$ segments of equal length., and
those ratings from the same segment will share the same rank. The
Netflix dataset has slightly over $100$ million ratings applied to
$17,770$ movies by $480,189$ users, where ratings are integers in
a $5$-star ordinal scale. 

\emph{Data likelihood estimation}. Table~\ref{tab:Average-log-likelihood}
shows the log-likelihood of test data averaged over
$100$ users with different numbers of movies per user. Results for
the Latent $\genModel$ are estimated using the AIS procedure.

\begin{table}
\begin{centering}
\begin{tabular}{l|c|c|c}
 & $M=10$ & $M=20$ & $M=30$\tabularnewline
\hline 
Plackett-Luce.MF & -14.7 & -41.3 & -72.6\tabularnewline
Latent $\genModel$ & -9.8 & -37.9 & -73.4\tabularnewline
\end{tabular}
\par\end{centering}

\caption{Average log-likelihood over $100$ users of test data (Movie Lens
$10$M dataset), after training on the $N=10$ movies per user ($K=10$).
$M$ is the number of test movies per user. The Plackett-Luce.MF and
the Latent OSM are comparable because they are both probabilistic
in ranks and can capture latent aspects of the data. The main difference
is that the Plackett-Luce.MF does not handle groupings or ties. \label{tab:Average-log-likelihood}}
\end{table}

\emph{Rank reconstruction}. Given the posterior vector, we ask whether
we can reconstruct the original rank of movies for that data instance.
For simplicity, we only wish to obtain a complete ranking, since it
is very efficient (e.g. a typical cost would be $N\log N$ per user).
Figure~\ref{fig:MovieLens-results}(c) indicates that high quality
rank reconstruction (on both training and test data) is possible given
enough hidden units. This suggests an interesting way to store and
process rank data by using vectorial representation.

\begin{figure}
\begin{centering}
\begin{tabular}{ccc}
\includegraphics[width=0.45\textwidth,height=0.4\textwidth]{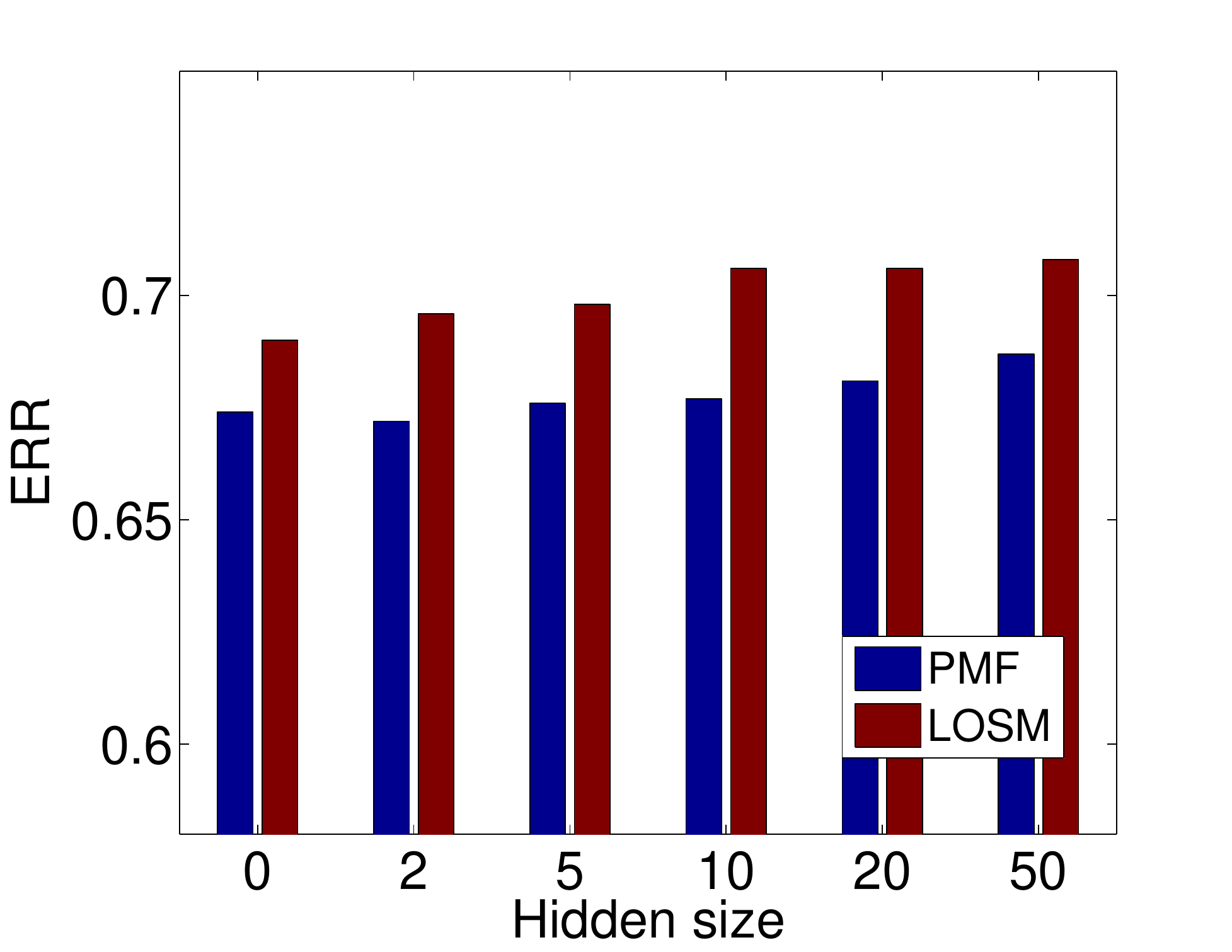} &  & \includegraphics[width=0.45\textwidth,height=0.4\textwidth]{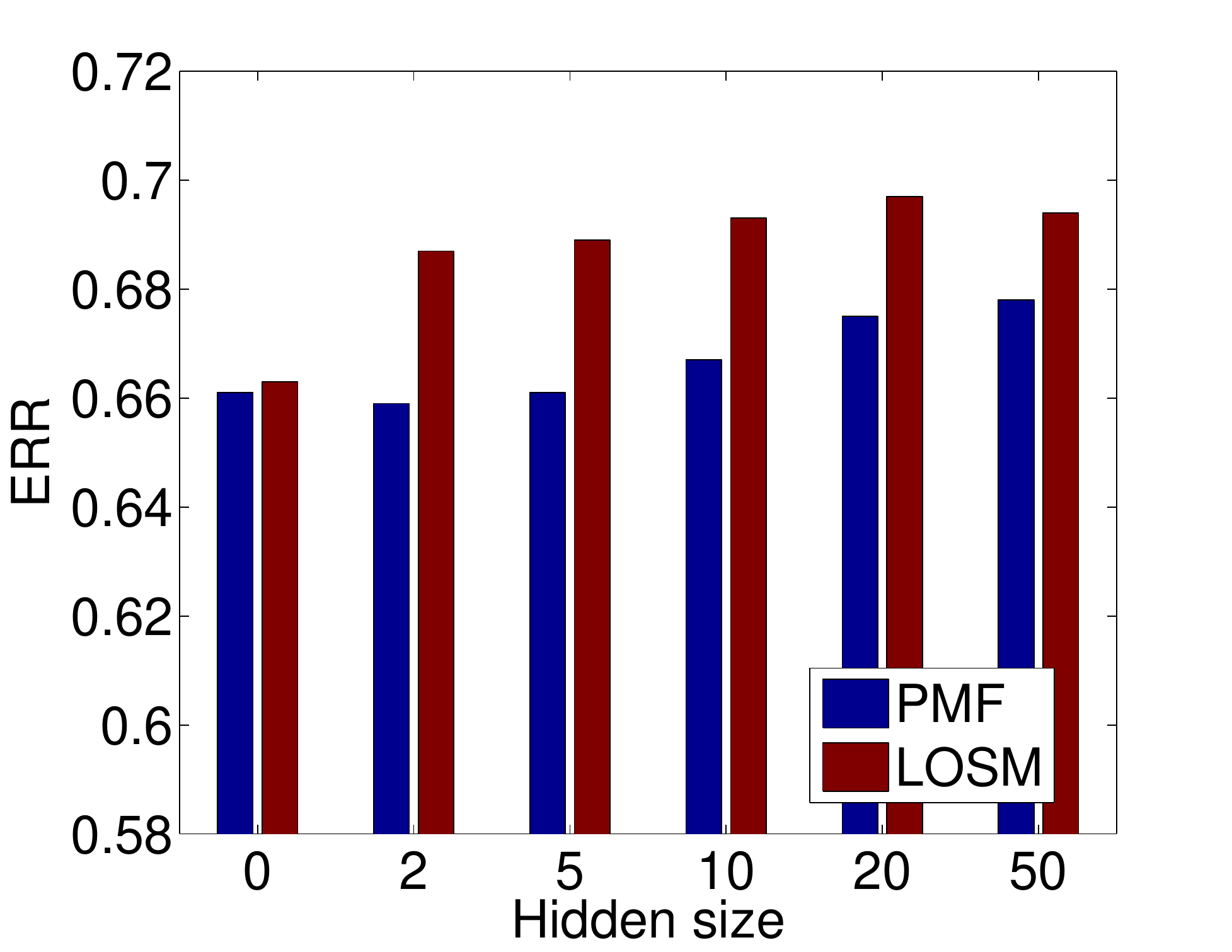}\tabularnewline
(a) MovieLens, $N=20$ &  & (b) Netflix $N=10$\tabularnewline
\end{tabular}
\par\end{centering}

\caption{Rank completion quality vs. number of hidden units - note that since
the PMF is not defined when hidden size is $0$, we substitute using
the result for hidden size $1$. \label{fig:ERR_vs_hidden_size}}
\end{figure}

\emph{Rank completion}. In collaborative filtering settings, we are
interested in ranking unseen movies for a given user. To highlight
the disparity between user tastes, we remove movies whose qualities
are inherently good or bad, that is when there is a general agreement
among users. More specifically, we compute the movie entropy as $H_{i}=-\sum_{r=1}^{5}P_{i}(r)\log P_{i}(r)$
where $P_{i}(r)$ is estimated as the proportion of users who rate
the movie $i$ by $r$ points. We then remove half of the movies with
lowest entropy. For each dataset, we split the data into a training
set and a test set as follows. For each user, we randomly choose $10$,
$20$ and $50$ items for training, and the rest for testing. To ensure
that each user has at least $10$ test items, we keep only those users
with no less than $20$, $30$ and $60$ ratings, respectively.

Figs.~\ref{fig:MovieLens-results}(b), \ref{fig:ERR_vs_hidden_size}(a)
and Table~\ref{tab:Results-MovieLens10M} report the results on the
MovieLens $10$M dataset; Figs.~\ref{fig:ERR_vs_hidden_size}(b)
and Table~\ref{tab:Results-Netflix100M} show the results for the
Netflix dataset. It can be seen that the Latent $\genModel$ performs
better than rivals when $N$ is moderate. For large $N$, the rating-based
method (PMF) seems to work better, possibly because converting rating
into ordering loses too much information in this case, and it is more
difficult for the Latent $\genModel$ to explore the hyper-exponential
state-space . 

\begin{table*}
\begin{centering}
\begin{tabular}{l|c|c|c|c|c|c}
 & \multicolumn{2}{c|}{$N=10$} & \multicolumn{2}{c|}{$N=20$} & \multicolumn{2}{c}{$N=50$}\tabularnewline
\cline{2-7} 
 & ERR & N@5 & {ERR} & {N@5} & {ERR} & {N@5}\tabularnewline
\hline 
{PMF} & 0.673 & 0.603 & 0.687 & 0.612 & \textbf{0.717} & \textbf{0.638}\tabularnewline
{pLPA} & 0.674 & 0.596 & 0.684 & 0.601 & 0.683 & 0.595\tabularnewline
{ListRank.MF} & 0.683 & 0.603 & 0.682 & 0.601 & 0.684 & 0.595\tabularnewline
{Plackett-Luce.MF} & 0.663 & 0.586 & 0.677 & 0.591 & 0.681 & 0.586\tabularnewline
{CoFi$^{RANK}$.Regress} & 0.675 & 0.597 & 0.681 & 0.598 & 0.667 & 0.572\tabularnewline
{CoFi$^{RANK}$.Ordinal} & 0.623 & 0.530 & 0.621 & 0.522 & 0.622 & 0.515\tabularnewline
{CoFi$^{RANK}$.N@10} & 0.615 & 0.522 & 0.623 & 0.517 & 0.602 & 0.491\tabularnewline
\hline 
\textbf{Latent }$\boldsymbol{\genModel}$ & \textbf{0.690} & \textbf{0.619} & \textbf{0.708} & \textbf{0.632} & 0.710 & 0.629\tabularnewline
\end{tabular}
\par\end{centering}

\caption{Model comparison on the MovieLens data for rank completion ($K=50$).
N@$T$ is a shorthand for NDCG@$T$. \label{tab:Results-MovieLens10M}}
\end{table*}

\begin{table*}
\begin{centering}
\begin{tabular}{l|c|c|c|c|c|c|c|c|}
 & \multicolumn{4}{c|}{{$N=10$}} & \multicolumn{4}{c|}{{$N=20$}}\tabularnewline
\cline{2-9} 
 & {ERR} & {N@1} & {N@5} & N@10 & {ERR} & {N@1} & {N@5} & N@10\tabularnewline
\hline 
{PMF} & 0.678 & 0.586 & 0.607 & 0.649 & 0.691 & 0.601 & 0.624 & 0.661\tabularnewline
{ListRank.MF} & 0.656 & 0.553 & 0.579 & 0.623 & 0.658 & 0.553 & 0.577 & 0.617\tabularnewline
\hline 
\textbf{Latent }$\boldsymbol{\genModel}$ & \textbf{0.694} & \textbf{0.611} & \textbf{0.628} & \textbf{0.666} & \textbf{0.714} & \textbf{0.638} & \textbf{0.648} & \textbf{0.680}\tabularnewline
\end{tabular}
\par\end{centering}

\caption{Model comparison on the Netflix data for rank completion ($K=50$).
\label{tab:Results-Netflix100M}}
\end{table*}

\section{Related Work \label{sec:Related-Work}}

This work is closely related to the emerging concept of \emph{preferences
over sets} in AI \cite{brafman2006preferences,wagstaff2010modelling}
and in social choice and utility theories \cite{barbera200417}. However,
most existing work has focused on representing preferences and computing
the optimal set under preference constraints \cite{binshtok2007computing}.
These differ from our goals to model a distribution over all possible
set orderings and to learn from example orderings. Learning from expressed
preferences has been studied intensively in AI and machine learning,
but they are often limited to pairwise preferences or complete ordering
\cite{cohen1999learning,weimer2008cofi}.

On the other hand, there has been very little work on learning from
ordered sets \cite{yue2008predicting,wagstaff2010modelling}. The
most recent and closest to our is the PMOP which models ordered sets
as a locally normalised \emph{high-order Markov chain} \cite{Truyen:2011a}.
This contrasts with our setting which involves
a globally normalised log-linear solution. Note that since the high-order
Markov chain involves all previously ranked subsets, while our OSM
involves pairwise comparisons, the former is not a special case of
ours. Our additional contribution is that we model the space of partitioning
and ordering directly and offer sampling tools to explore the space.
This ease of inference is not readily available for the PMOP. Finally,
our solution easily leads to the introduction of latent variables,
while their approach lacks that capacity.

Our \texttt{\small split-and-merge} sampling procedure bears some
similarity to the one proposed in \cite{jain2004split} for mixture
assignment. The main difference is that we need to handle the extra
orderings between partitions, while it is assumed to be exchangeable
in \cite{jain2004split}. This causes a subtle difference in generating
proposal moves. Likewise, a similar method is employed in \cite{ranganathan2006bayesian}
for mapping a set of observations into a set of landmarks, but again,
ranking is not considered.

With respect to collaborative ranking, there has been work focusing
on producing a set of items instead of just ranking individual ones
\cite{price2005optimal}. These can be
considered as a special case of $\genModel$ where there are only
two subsets (those selected and the rest).

\section{Conclusion and Future Work}

We have introduced a latent variable approach to modelling ranked
groups.\emph{ }Our main contribution is an efficient \texttt{\small split-and-merge}
MCMC inference procedure that can effectively explore the hyper-exponential
state-space.\emph{ }We demonstrate how the proposed model can be useful
in collaborative filtering. The empirical results suggest that proposed
model is competitive against state-of-the-art rivals on a number of
large-scale collaborative filtering datasets.

\end{document}